\documentclass{elsarticle}

\makeatletter
\def\ps@pprintTitle{%
\let\@oddhead\@empty
\let\@evenhead\@empty
\def\@oddfoot{}%
\let\@evenfoot\@oddfoot}
\makeatother

\usepackage{lineno,hyperref}
\usepackage{amsmath}
\usepackage{rotating}
\usepackage{amsfonts}
\usepackage{subcaption}
\usepackage{amstext}
\usepackage{algorithm}
\usepackage{algorithmic}
\usepackage{amssymb}
\usepackage{color}
\modulolinenumbers[1]
\definecolor{nbarrier}{RGB}{255, 120, 50}
\definecolor{nbicycle}{RGB}{255, 192, 203}
\definecolor{nbus}{RGB}{255, 255, 0}
\definecolor{ncar}{RGB}{0, 150, 245}
\definecolor{nconstruct}{RGB}{0, 255, 255}
\definecolor{nmotor}{RGB}{200, 180, 0}
\definecolor{npedestrian}{RGB}{255, 0, 0}
\definecolor{ntraffic}{RGB}{255, 240, 150}
\definecolor{ntrailer}{RGB}{135, 60, 0}
\definecolor{ntruck}{RGB}{160, 32, 240}
\definecolor{ndriveable}{RGB}{255, 0, 255}
\definecolor{nother}{RGB}{139, 137, 137}
\definecolor{nsidewalk}{RGB}{75, 0, 75}
\definecolor{nterrain}{RGB}{150, 240, 80}
\definecolor{nmanmade}{RGB}{213, 213, 213}
\definecolor{nvegetation}{RGB}{0, 175, 0}
\usepackage[table]{xcolor}
\usepackage{xcolor}

\begin{document}
\begin{frontmatter}

\title{FSF-Net: Enhance 4D Occupancy Forecasting \\with Coarse BEV Scene Flow for\\ Autonomous Driving}

\author{Erxin Guo$^{1}$, Pei An$^{1}$, You Yang$^{1}$, Qiong Liu$^{1,*}$, and An-An Liu$^{2}$}
\address{1. School of Electronic
Information and Communications,\\ Huazhong University of Science and Technology, Wuhan 430074, China.}
\cortext[mycorrespondingauthor]{Corresponding author: q.liu@hust.edu.cn}
\address{2. School of Electrical and Information Engineering, \\Tianjin University,Tianjin 300072, China}

\begin{abstract}

4D occupancy forecasting is one of the important techniques for autonomous driving, which can avoid potential risk in the complex traffic scenes. Scene flow is a crucial element to describe 4D occupancy map tendency. However, an accurate scene flow is difficult to predict in the real scene. In this paper, we find that BEV scene flow can approximately represent 3D scene flow in most traffic scenes. And coarse BEV scene flow is easy to generate. Under this thought, we propose 4D occupancy forecasting method FSF-Net based on coarse BEV scene flow. At first, we develop a general occupancy forecasting architecture based on coarse BEV scene flow. Then, to further enhance 4D occupancy feature representation ability, we propose a vector quantized based Mamba (VQ-Mamba) network to mine spatial-temporal structural scene feature. After that, to effectively fuse coarse occupancy maps forecasted from BEV scene flow and latent features, we design a U-Net based quality fusion (UQF) network to generate the fine-grained forecasting result. Extensive experiments are conducted on public Occ3D dataset. FSF-Net has achieved IoU and mIoU 9.56\% and 10.87\% higher than state-of-the-art method. Hence, we believe that proposed FSF-Net benefits to the safety of autonomous driving.  

\end{abstract}


\end{frontmatter}


\section{Introduction}


Four-dimensional (4D) occupancy forecasting is one of the important techniques for autonomous driving. It can forecast the future occupancy map from  history data. This task is able to determine the trajectories of fast-moving objects, which can avoid potential risk in the complex traffic scene \cite{occ-world}. In general, occupancy prediction and occupancy forecasting are sister tasks related to occupancy map in the context of autonomous driving \cite{occ-world, Occ3d-dataset}. Occupancy prediction is to generate occupancy map from the imaging sensors, such as camera (monocular and camera array) \cite{occ-pred-1, occ-pred-2}, light detection and ranging (LiDAR) \cite{esc-net, lidar-occ-1, Cls}, and the combination of camera and LiDAR (or say, LiDAR-camera system) \cite{multi-sensor-occ, multi-sensor-occ-2, multi-sensor-occ-3, multi-sensor-occ-4}. While occupancy forecasting is to estimate occupancy map in the future, with the inputs of a series of history occupancy maps \cite{occ-world}. Both of them can reduce the traffic risk sharply for autonomous driving. 


However, 4D occupancy forecasting is more difficult than its 3D occupancy prediction. Its challenge is \textit{how to efficiently learn
the tendency of 4D occupancy maps}. In the current stage, most researchers focuses on learning spatial feature from sensor data for occupancy prediction \cite{lidar-occ-1}. While, for occupancy forecasting, this task needs to abstract 4D spatial and temporal occupancy features. But, the occupancy feature map with the additional temporal dimension hinders the learning efficiency, for the existing 2D or 3D convolution layers based network is difficult to mine both temporal and spatial features from history occupancy maps.



To deal with the above challenge, current researchers attempt to learn 4D occupancy map tendency in a manner of latent feature reconstruction \cite{occ-world}. This kind of method consists of two steps. In the first step, it encodes and learns latent spatio-temporal features from historical occupancy maps. In the second step, it predicts the latent features based on the encoded historical spatio-temporal features and reconstructs the future 3D occupancy map from the predicted latent features. 
This two-stage 4D occupancy forecasting method is more likely to artificial intelligence generated content (AIGC) \cite{chatgpt}. In 2024, Zheng et al. developed a state-of-the-art work named as OccWorld \cite{occ-world}, for 4D occupancy forecasting. In their scheme, they leveraged a generative pre-training transformers (GPT) \cite{chatgpt} to memorize the temporal occupancy features and reconstruct future occupancy maps. Although OccWorld outperforms other approaches in the public nuScenes \cite{nuscenes} and Occ3D \cite{Occ3d-dataset} datasets, the forecasting metrics, such as interaction-over-union (IoU) of occupancy labels and mean IoU (mIoU) of semantic labels, still have the large improvement space. 

Recently, many researchers notice that scene flow is an important element in occupancy forecasting \cite{forecast-0}. However, it is still difficult to predict an accurate 3D scene flow from the complex traffic scene. Also, scene flow prediction adds the computation burden.  Hence, it is still essential to explore a more efficient way to represent 4D occupancy map tendency. 

To deal with above problem, we find that bird's eye view (BEV) scene flow might be the key of occupancy forecasting, for (i) BEV scene flow can approximately describe 3D scene flow in most traffic scenes \cite{BEV} and (ii) coarse BEV scene flow is easy to obtain without deep learning. Under this thought, we propose a occupancy forecasting network FSF-Net based on coarse BEV scene flow. At first,  we develop a general occupancy forecasting architecture based on coarse BEV scene flow. This architecture can forecast the occupancy map in a coarse-to-fine manner. And we provide a fast trick to estimate BEV scene flow. Then, to further enhance the representation ability of history occupancy feature, we design a vector quantized
based Mamba (VQ-Mamba) network to abstract the salient spatial-temporal structural scene feature. After that, to effectively fuse coarse occupancy maps forecasted from scene flow and
latent features, we design a U-Net based quality fusion (UQF)
network to generate the fine-grained high-quality forecasting result. 

In the last, we conduct extensive experiment on the public Occ3D dataset \cite{Occ3d-dataset}. FSF-Net has achieved IoU and mIoU 9.56\%
and 10.87\% higher than state-of-the-art method \cite{occ-world}. Hence, we believe that the proposed method benefits to the safe autonomous driving. 


Our core contribution is \textit{leveraging the imperfect BEV scene flow as auxiliary feature to enhance learning efficiency of 4D occupancy map tendency}. Based on this thought, we propose a FSF-Net to effectively fuse latent occupancy feature and BEV scene flow for 4D occupancy forecasting task. 

\section{Related Works}

In the current stage, there are fewer published works about occupancy forecasting than occupancy prediction \cite{occ-world}. Existing occupancy forecasting method has the close relation to occupancy prediction, for both of them have the highly similar network architecture to encode occupancy feature. So, we illustrate the mainstream approaches of occupancy prediction and occupancy forecasting in this related works. 

Occupancy prediction task is first discussed. It aims to generate occupancy map of the surrounding using sensor data. If input is sparse LiDAR point cloud, this task is highly similar to point cloud completion\cite{Complete, Complete-1}. Wang et al. \cite{occ-op-nav} leveraged a U-Net based 3D occupancy prediction network to enhance the safety navigation of unmanned aerial vehicle (UAV). Jiang et al. \cite{Dynamic} enhanced the representation of detailed parts using dynamic graph convolutional occupancy networks. Xia et al. \cite{scp-net} leveraged a teacher-student based knowledge distillation network for occupancy prediction. Teacher model predicts 3D occupancy map from multi-frame unlabeled LiDAR frames. The predicted map can be regarded as pseudo label and can be used to supervise  student model. Kung et al. \cite{radar-occ} developed a Radar based 2D occupancy prediction network. As Radar based occupancy GT label is rare, they collected labels with BEV map projected by LiDAR point cloud. Rather than using unlabeled data, An et al. \cite{esc-net} analyzed that LiDAR based 3D occupancy prediction suffers from triplet sparsity, such as input sparsity, foreground object sparsity, and ground truth (GT) sparsity. They designed a sparse and dense 3D convolution based U-Net with using feature-level completion. Inspired by previous work \cite{occ-op-nav}, Vizzo et al. \cite{tsdf-occ} developed a U-Net based completion network to complete the sparse truncated signed distance (TSDF) map generated by raw LiDAR point cloud. Other researchers focus on the downstream application of occupancy prediction. Wiesmann et al. \cite{loc-ndf} enhance the efficiency of robot localization with the usage of occupancy neural field.

In the meanwhile, some of researchers study 3D occupancy prediction based on LiDAR-camera system\cite{Lidar-camera}. Wu et al. \cite{multi-sensor-occ-2} noticed that a dense predicted occupancy map is beneficial to 3D object detection. They used image based LiDAR depth completion to generate the dense 3D occupancy map. In 2024, Pan et al. \cite{multi-sensor-occ} attempted to enhance the accuracy of occupancy prediction via a geometric and semantic based multi-modality feature fusion.

Although LiDAR-camera based occupancy prediction can achieve the most accurate performance, researchers tend to study camera based occupancy prediction, for camera (no matter for monocular camera or multi-camera system) is much cheaper than LiDAR sensor, And the sparsity of the point cloud hinders the extraction of scene representation \cite{Low}. Kundu et al. \cite{panp-nf} designed a deep neural network (DNN) to predict a panoptic neural field from a series of panoptic images. This neural fields can generate occupancy map with the semantic and instance labels. Cao et al. \cite{mono-scene} developed an occupancy prediction network with combining both 2D and 3D U-Nets. They designed a module of Features Line of Sight Projection (MLSP) as a bridge to convert 2D feature map as 3D feature map. Recently, Huang et al. \cite{occ-pred-1} presented a DNN architecture TVPFormer to reconstruct 3D occupancy map from the surround-view cameras. They were the first to mine spatial feature from Tri-perspective view (TPV) rather than BEV view, thus enhancing the quality of occupancy prediction. Instead of designing new DNN, Wei et al. \cite{occ-pred-2} reconstructed the dense occupancy map with time-continuous LiDAR point clouds. This scheme enriches GT labels for 3D occupancy prediction task. To enhance the occupancy prediction accuracy from single-view, Zhang et al. \cite{occ-former} considered the close relation of semantic label and occupancy label, and designed a transformer based dual-path architecture to augment semantic and occupancy features.

From the above discussions, it is obvious that 3D occupancy prediction has been widely studied these years. However, the study of 4D occupancy forecasting has just begun. In the beginning stage, mainstream methods is to forecast the motion tendency of objects, such as vehicles and pedestrian \cite{forecast-0, Sceneflow-1}. Mahjourian et al. \cite{forecast-0} found that scene flow has an important role in forecasting object-level occupancy map. But, the forecasting quality depends on scene flow accuracy. Now, more and more researchers are interested in scene-level occupancy forecasting, for it has wide effect in autonomous driving. Khurana et al. \cite{forecast-1} focused on 3D free-space forecasting from time-continuous LiDAR scans. Ferenczi et al. \cite{forecast-2} were the first to design end-to-end network for occupancy forecasting. But, their scheme requires road-graph and agent information, which is inconvenient to use in the actual scene. Based on their previous work \cite{forecast-1}, Khurana et al. \cite{forecast-3} proposed a differential render to supervise the occupancy maps forecasting with LiDAR point clouds. With this render, they designed a neural network to both forecast LiDAR point cloud and 3D occupancy map. In 2024, Zheng et al. \cite{occ-world} proposed a GPT based world model, OccWorld. It had achieved the best performance in 4D occupancy forecasting. Its core is to use GPT architecture \cite{chatgpt} to mine 4D occupancy tendency. However, their IoU and mIoU metric performances sill have the large improvement. 

Until now, although 4D occupancy forecasting has made some progress, it still lacks an efficient framework to ensure the generalization ability of 4D occupancy forecasting in the open traffic scene. Some researchers, like Mahjourian et al. \cite{forecast-0}, noticed the importance of scene flow to occupancy forecasting task. However, they does not fully consider the noise of predicted scene flow. To deal with this issue, we propose BEV scene flow based method, FSF-Net, in the following.

\section{Methodology}
\begin{figure*}[t]
	\centering
		\includegraphics[width=1.0\linewidth]{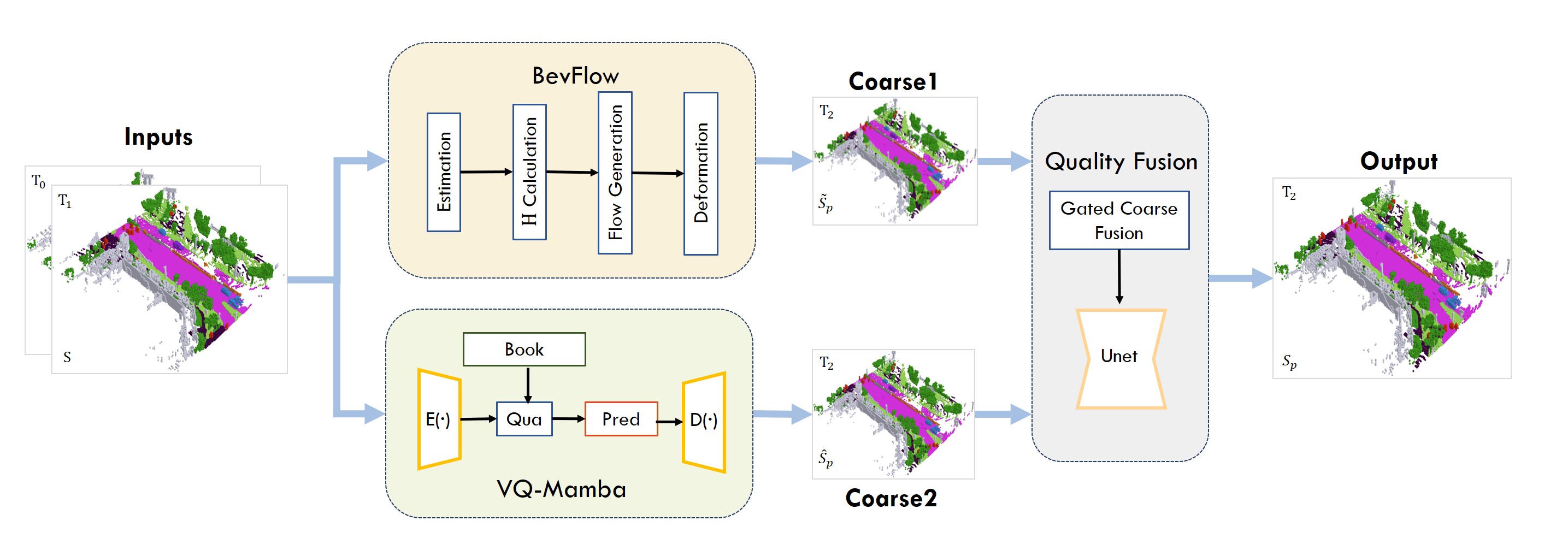}	
		\caption{The BEV Flow module and VQ-Mamba module achieve coarse scene prediction through voxel movement relationships and neural networks, respectively. Then, the results are aggregated through the Quality-Fusion module to obtain the final detailed prediction outcome.}
\label{fig:Framework}
\end{figure*}

\begin{figure*}[t]
	\centering
		\includegraphics[width=1.0\linewidth]{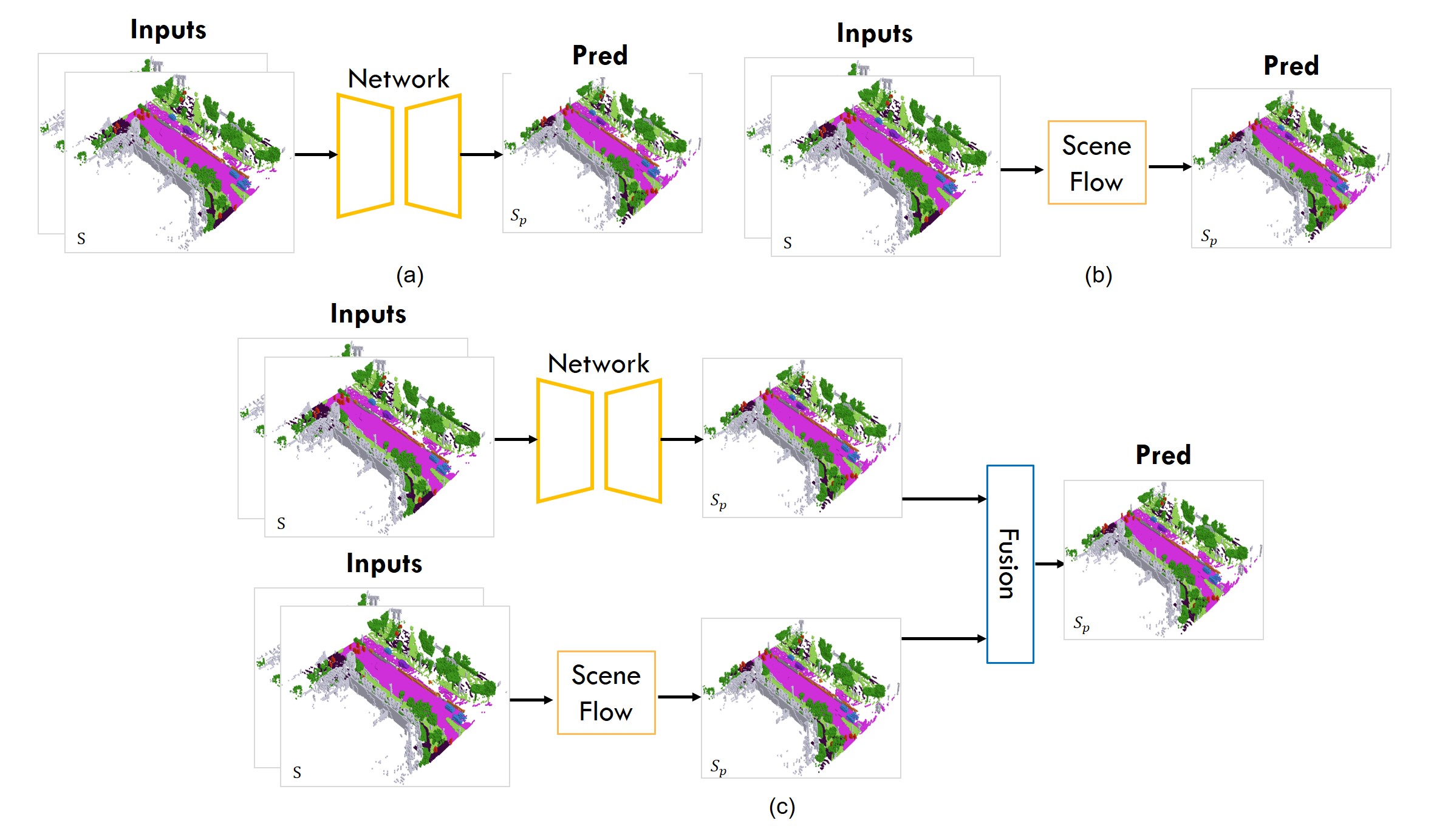}	
		\caption{(a) Framework of mainstream methods (b) Framework based on scene flow \\
        (c) Framework of fusion method}
\label{fig:process}
\end{figure*}
\begin{figure*}[t]
	\centering
		\includegraphics[width=1.0\linewidth]{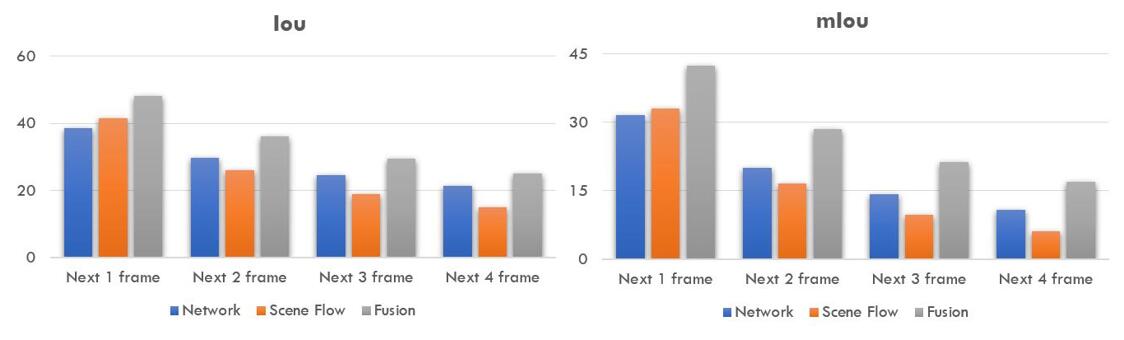}	
		\caption{Late Fusion can combine the advantages of both prediction methods to achieve performance improvements.}
\label{fig:compare}
\end{figure*}
This section elaborates on the 4D occupancy forecasting method in detail. Firstly, the 4D occupancy forecasting problem is defined. Then, the overall network framework of the proposed 4D occupancy forecasting method is introduced. Finally, the key components of each module in the proposed 4D occupancy forecasting method are presented.
\subsection{Problem definition}

Given a scene sequence containing n consecutive frames $\mathcal{S} = \{\mathcal{S}^{0},\mathcal{S}^{1},\ldots,\mathcal{S}^{n}\}$ as input,where the dimensions of each frame are $(\mathcal{H},\mathcal{W},\mathcal{L})$,representing height,width,and depth,and $\mathcal{S}(h,w,l)$ denotes the semantic label of the LiDAR voxel at coordinates $(h,w,l)$.If the voxel at $(h, w, l)$ is unoccupied, its value is $0$.The goal of 4D occupancy forecasting is to output the sequence of $m$ consecutive frames corresponding to the future scene $\mathcal{S}_p = \{\mathcal{S}^{n+1},\mathcal{S}^{n+2},\ldots,\mathcal{S}^{n+m}\}$.Typically, 4D 4D occupancy forecasting operates under the condition of having ground truth. During the training phase, the ground truth sequence of the future $m$ frames is provided, and the network's training results are compared with the ground truth. In the testing phase, only the input sequence of the scene is provided, and predictions are made directly.

\subsection{BEV Scene Flow Based forecasting Framework}
From Sec.1, we can see that the process of current mainstream techniques for 4D occupancy forecasting tasks is shown in Fig. \ref{fig:process}(a). The core of these techniques is to learn the latent features of the current occupancy scene and predict the latent features of the next frame’s occupancy scene using neural networks. Since neural networks aggregate temporal information through deep learning models, errors such as incorrect predictions of moving object positions, object deformation, and semantic label errors may occur when predicting scene details. Considering that scene flow contains information about the movement of objects in the scene, it can better represent details in future scenes. Therefore, as shown in Fig. \ref{fig:process}(b), we consider using scene flow to deform the current frame to obtain the prediction results of future scenes. However, since scene flow only includes the movement information of objects in the current frame, predicting future scenes based solely on scene flow performs well when predicting occupancy grids that are present in both the current and future scenes but performs poorly when predicting occupancy grids that exist in future scenes but not in the current scene. Therefore, as shown in Fig. \ref{fig:process}(c), we aim to integrate these two approaches to retain the advantages of both methods and achieve more precise 4D occupancy forecasting. Additionally, considering the difficulty of fine-grained scene flow prediction and the fact that fine-grained scene flows are generally predicted using complex neural networks, which often leads to longer prediction times, we have designed a coarse scene flow from a BEV perspective. This approach aims to obtain scene flow using only physical information without neural networks. Since the introduced scene flow is relatively coarse and not easily aligned with the occupancy scene, directly performing early fusion might lead to a decline in prediction quality. Therefore, we adopted a late fusion approach to combine the prediction results of the two methods. The aim is for our designed fusion framework to successfully integrate the advantages of both methods, achieving the performance improvement as shown in Fig.\ref{fig:compare}.

\begin{figure*}[t]
	\centering
		\includegraphics[width=1.0\linewidth]{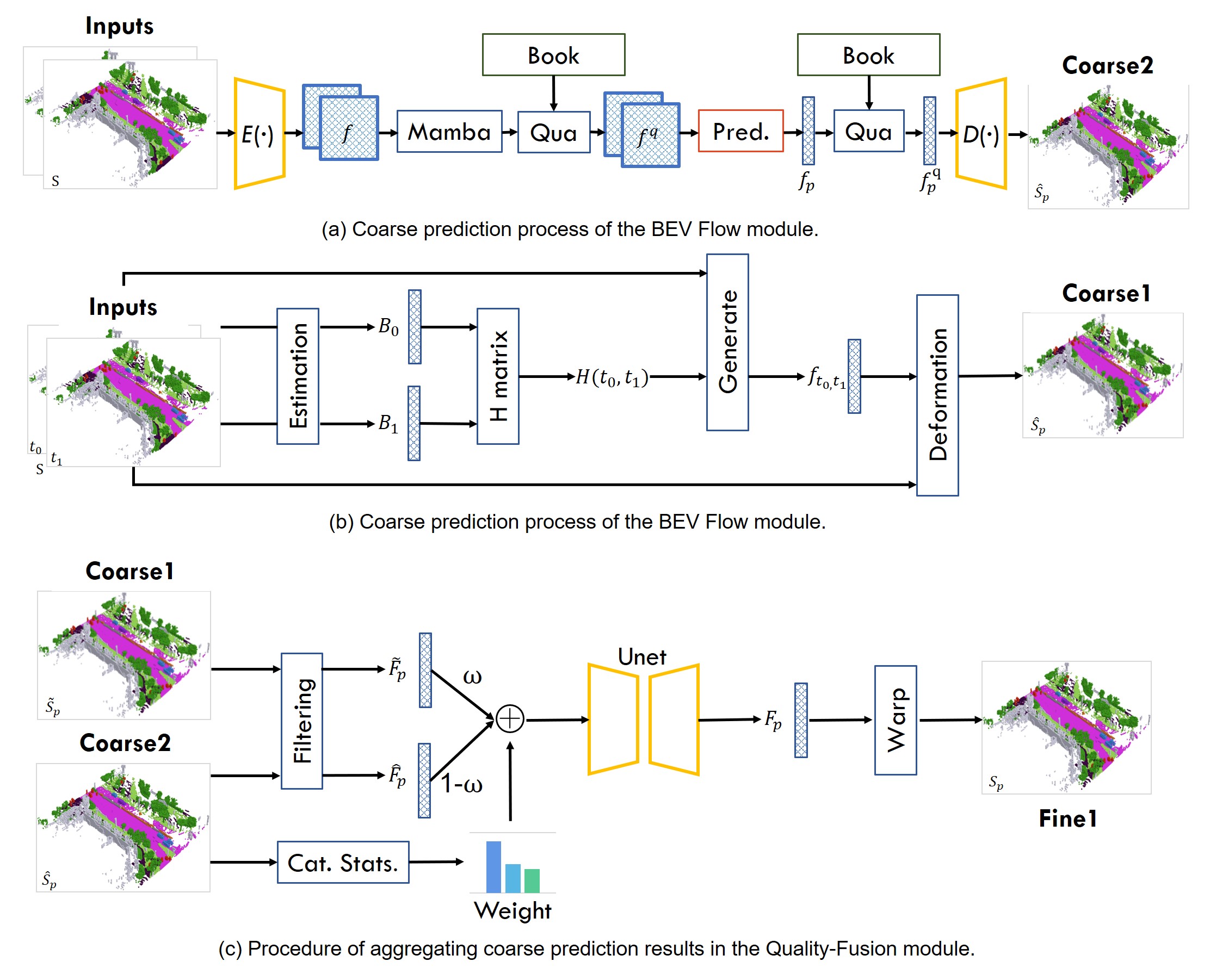}	
		\caption{Composition of the prediction network.}
\label{fig:module}
\end{figure*}

As shown in Fig. \ref{fig:Framework}, the forecasting Framework mainly includes a BEV Flow module(See in Fig. \ref{fig:module}, Appendix A.), a VQ-Mamba module, and a Quality-Fusion module. The first two can perform coarse predictions on subsequent multi-frame scenarios. The prediction network first passes the input voxel scene to both the BEV Flow module and the VQ-Mamba module to obtain the predicted scene.The BEV Flow module predicts future scene occupancy information by leveraging scene flow in the BEV space. Meanwhile, the VQ-Mamba module learns the latent features of the scene based on VQVAE and predicts future scene occupancy information using the prediction module. Additionally, to enable the model to better capture temporal information in scene changes, we integrate the Mamba module into the VQ-Mamba module. After obtaining the prediction results from BEV Flow and VQ-Mamba, we use the Quality-Fusion module to fuse the previously obtained coarse prediction results, aiming to retain the advantages of the predictions from the BEV Flow and VQ-Mamba modules, resulting in a more refined scene prediction outcome.The overall forecasting framework can be expressed as follows.
\begin{equation}
    S_p = P_{QF}\left( P_{BEV}(S), P_{VQ}(S) \right)
\end{equation}
where $S$  represents the occupancy scene at the current time step.$S_p$ represents the predicted scene at a future time step. $P_{BEV}(S)$ denotes the prediction of the scene using the BEV Flow module. $P_{VQ}(S)$ represents the prediction of the scene using the VQ-Mamba module. $P_{QF}(P_{BEV}, P_{VQ})$ denotes the result of fusing both predictions using the Quality-Fusion module. In the next subsections, this paper will subsequently depict the VQ-Mamba module, and the Quality-Fusion module.

\subsection{VQ-Mamba}

The Mamba module benefits from its built-in causal convolution component, which endows it with strong temporal information extraction capabilities. Since temporal information is crucial for scene prediction tasks, we introduced Mamba into the prediction network and proposed a two-stage VQ-Mamba network, as shown in Fig. \ref{fig:module}(b).\par
In the first stage, to enable the codebook to learn scene features, following the approach of most previous work, we first use an encoder to extract features $\mathcal{F}$ from the input scene $\mathcal{S}$.Next, we input the features $\mathcal{F}$ into the codebook to enable it to learn and reconstruct scene characteristics.Besides, To better leverage the temporal correlations between input frames, we incorporated the Mamba structure before the codebook, aiming to improve the learning of scene features through Mamba. The process of learning the codebook can be expressed as follows.
\begin{equation}
f^{q} = \min_{c \in C} \|\text{Mamba}(\hat{f}^{q}) - c\|_2^2
\end{equation}
where $f^{q}$ represents the discrete scene feature, C corresponds to the codebook, and c denotes the code in the codebook. Finally, we restore the reconstructed scene features $\mathcal{F}$ to the original scene $\mathcal{S}$ using a decoder. In the second stage, after obtaining the scene features $\mathcal{F}$ from the encoder, we retrieve the corresponding codebook indices for the current scene features from the codebook. The process of obtaining the voxel indices for the current frame can be expressed as follows.
\begin{equation}
\text{Index} = \text{min}_c \|\text{Mamba}(f^{q}) - c \|
\end{equation}
where $\text{Index}$ is the codebook index corresponding to the scene feature $f^{q}$.Then, using the U-Mamba module, we convert the current frame's scene feature $\text{Index}$ indices into indices for the future frame $\text{Index}_p$. Next, by querying the codebook for the scene features corresponding to these future frame indices, we obtain the scene features $\mathcal{F}_p^{q}$ for the future frame $\mathcal{S}_p$.
\begin{equation}
\mathcal{F}_p^{q} = \text{Codebook} \left( \text{U-Mamba}(\text{Index}_p) \right)
\end{equation}
where $\text{Codebook}$ represents the codebook query process, and $\text{U-Mamba}$ denotes the index prediction process.Finally, we use a decoder to convert the predicted future frame scene features $\mathcal{F}^{p}$ into a coarse prediction result $\hat{\mathcal{S}_p}$.

\subsection{Quality-Fusion}
To combine the strengths of the coarse prediction results $\widetilde{\mathcal{S}}_p$ and $\hat{\mathcal{S}_p}$
generated by the BEV Flow module and the VQ-Mamba module, we designed a Quality-Fusion module. The details are shown in Fig. \ref{fig:module}(c).\par
We divided the Quality-Fusion module into two parts: Gated Coarse Fusion and UNet Fine Fusion. In the Gated Coarse Fusion part, we use the filtering module to extract the most probable semantic information $\widetilde{\mathcal{F}}_p$, $\hat{\mathcal{F}_p}$ for each voxel location from the coarse matching results $\widetilde{\mathcal{S}}_p$, $\hat{\mathcal{S}_p}$. Then, we utilize a gating mechanism to fuse the filtered semantic information together.The gating and feature filtering process can be expressed as follows.
\begin{equation}
\mathbf{F}_p = (1 - w) \cdot \text{warp}(\arg\max(\widetilde{\mathcal{S}}_p
)) + w \cdot \text{warp}(\arg\max(\hat{\mathcal{S}_p}))
\end{equation}
Where $w$ represents the gating weight, and $\mathbf{F}_p$ denotes the scene features obtained from the initial fusion of the coarse prediction scene $\widetilde{\mathcal{S}}_p$ and $\hat{\mathcal{S}_p}$.After gating weight fusion, the semantic information from the same locations in $\widetilde{\mathcal{F}}_p$, $\hat{\mathcal{F}_p}$ is mapped to the corresponding position in the same scene feature $\mathbf{F}_p$. As a result, the same location in $\mathbf{F}_p$ often contains semantic information from two different classes, requiring further fusion. Considering that different classes appear with varying frequencies in the scene, we designed class weights based on the occurrence frequency of each class in the coarse prediction result $\hat{\mathcal{S}_p}$ and incorporated these weights into the predicted feature $\mathbf{F}_p$ to achieve better fusion of the semantic information from different classes at the same location.The process of generating and incorporating the weights is as follows.

\begin{equation}
\alpha[i] = \text{sorting}\left(\sum_{j=1}^{n} \delta\left(\hat{\mathcal{S}_p} = i\right)\right) \quad \text{for } i \in [0, 17]
\end{equation}
\vspace{-0.5cm}
\begin{equation}
\mathbf{F}_p^T = \alpha \otimes \mathbf{F}_p
\end{equation}
Where $\alpha[i]$ represent the weight corresponding to the 
i-th category,with a total of 18 categories.And $\mathbf{F}_p^T$ denotes the scene features with added category weights.Finally,  in the UNet Fine Fusion part, we input the scene features  $\mathbf{F}_p^T$ 
into a U-Net network to achieve semantic fusion at different resolutions. The fused results are then filtered to obtain the final refined scene prediction.The fusion process of the UNet module is represented as follows.
\begin{equation}
\mathcal{S}_p = \text{argmax}\left( \text{U-Net}\left(\mathbf{F}_p^T \right)\right)
\end{equation}
Where $\mathcal{S}_p$ represent the final refined scene prediction.For the Quality-Fusion module, we use the following formula for training.
\begin{equation}
\mathbf{Loss} = \mathcal{L}_{\text{soft}}(P_{QF}\left(\widetilde{\mathcal{S}}_p, \hat{\mathcal{S}_p}\right),S_p)+ \lambda \mathcal{L}_{\text{lovasz}}(P_{QF}\left(\widetilde{\mathcal{S}}_p, \hat{\mathcal{S}_p}\right), S_p)
\end{equation}
where  $\mathcal{L}_{\text{soft}}$ represents the softmax loss. $\mathcal{L}_{\text{lovasz}}$ represents the Lovasz-softmax loss. $S_p$ denotes the ground truth labels. $\lambda$ is a balance factor.

\section{Experiments}

\subsection{Implementation details}
To evaluate the effectiveness of our FSF-Net, we conducted comprehensive experiments on public Occ3D dataset\cite{Occ3d-dataset} which serves as a benchmark for evaluating tasks related to 3D occupancy prediction. The dataset is derived from the nuScenes dataset\cite{nuscenes} and aims to assist tasks related to 3D occupancy prediction by providing dense 3D occupancy ground truth. The dataset contains a total of 40,000 frames, with 600 scenes used for training, 150 scenes for validation, and an additional 150 scenes for testing. It covers 16 common categories, as well as a generic object (GO) category. Each sample spans a spatial range from [40m,40m,-1m] to [40m,40m,5.4m]. In the 4D occupancy forecasting task, we use mIoU and IoU as evaluation metrics to assess the accuracy of predicting future 3D occupancy results. Given that BEV images are frequently used in autonomous driving, we also  assess the prediction results using mIoU and IoU metrics in the BEV space. Additionally, we use L2 error and collision rate as evaluation metrics to measure the discrepancy between the planned and actual trajectories, as well as the probability of potential collisions.\par
Next, we provide the details of the proposed FSF-Net network.  we use 4 historical frames to predict the subsequent 4 future frames. The encoder in the VQ-Mamba component uses a down-sampling factor of 4. The subsequent codebook consists of 512 nodes with a 128-dimensional feature representation. The prediction part also employs a 4x down-sampling factor and includes 3 scales. The Quality-Fusion module also adopts a 4x down-sampling factor and performs four down-sampling steps.During training, the learning rate and optimizer were kept as the default settings from previous work, with a batch size of 1. All experiments were conducted on a single NVIDIA GeForce RTX 4080 GPU.

\subsection{Evaluation of 4D Occupancy Forecasting}
We compared our model with three methods: (1) OccWorld\cite{occ-world}: A paradigm that simultaneously predicts the evolution of surrounding scenes and plans future trajectories for autonomous vehicles. It uses multi-scale perception and cross-attention to predict future scene labels and self-labels, enabling future occupancy prediction. (2) BEV-Flow: A non-neural network-based prediction method that predicts future occupancy by modeling the scene flow between consecutive frames. (3) Base-Net: A foundational occupancy prediction network based on our proposed coarse-to-fine prediction framework. It employs a cross-attention mechanism for coarse-to-fine fusion to achieve high-precision occupancy prediction results.\par
Additionally, inspired by OccWorld, we designed a Copy\&Paste method for comparison. This method copies the current ground-truth occupancy data as future observations, aiming to assess whether our model can learn the underlying scene evolution capabilities

\begin{table*}[!ht]
\centering
\caption{Performance of 4D occupancy forecasting in Occ3D validation dataset. Term Next-$A$ means forecasting the next $A$-th frame.}
\resizebox{1.0\linewidth}{!}{
\rotatebox{0}{
\begin{tabular}{c| c | c c | c c c c c c c c c c c c c c c c}
    \hline
    \textbf{Method}
    & {Frame} 
    & IoU& mIoU
    & \rotatebox{90}{\textcolor{nbarrier}{$\blacksquare$} barrier}
    & \rotatebox{90}{\textcolor{nbicycle}{$\blacksquare$} bicycle}
    & \rotatebox{90}{\textcolor{nbus}{$\blacksquare$} bus}
    & \rotatebox{90}{\textcolor{ncar}{$\blacksquare$} car}
    & \rotatebox{90}{\textcolor{nconstruct}{$\blacksquare$} const. veh.}
    & \rotatebox{90}{\textcolor{nmotor}{$\blacksquare$} motorcycle}
    & \rotatebox{90}{\textcolor{npedestrian}{$\blacksquare$} pedestrian}
    & \rotatebox{90}{\textcolor{ntraffic}{$\blacksquare$} traffic cone}
    & \rotatebox{90}{\textcolor{ntrailer}{$\blacksquare$} trailer}
    & \rotatebox{90}{\textcolor{ntruck}{$\blacksquare$} truck}
    & \rotatebox{90}{\textcolor{ndriveable}{$\blacksquare$} drive. suf.}
    & \rotatebox{90}{\textcolor{nother}{$\blacksquare$} other flat}
    & \rotatebox{90}{\textcolor{nsidewalk}{$\blacksquare$} sidewalk}
    & \rotatebox{90}{\textcolor{nterrain}{$\blacksquare$} terrain}
    & \rotatebox{90}{\textcolor{nmanmade}{$\blacksquare$} manmade}
    & \rotatebox{90}{\textcolor{nvegetation}{$\blacksquare$} vegetation} \\
    \hline
Copy-Paste & Next-1 & 30.16 & 21.20 & 23.34 & 8.11 & 31.62 & 20.98 & 17.67 & 7.25 & 7.11 & 8.54 & 19.12 & 20.94 & 48.96 & 32.80 & 32.30 & 30.47 & 18.28 & 18.54 \\
OccWorld \cite{occ-world} & Next-1 & 38.59 & 31.51 & 37.53 & 22.28 & 38.01 & 37.29 & 28.76 & 20.34 & 30.06 & 20.26 & 24.85 & 36.16 & 46.95 & 37.28 & 36.80 & 32.17 & 29.26 & 32.65 \\
BEV-Flow & Next-1 & 41.51 & 32.98 & 35.81 & 22.12 & 34.87 & 27.66 & 38.89 & 18.64 & 21.25 & 22.15 & 30.45 & 34.15 & 53.77 & 40.88 & 43.13 & 40.25 & 29.60 & 37.78 \\
Base-Net & Next-1 & 39.67 & 34.10 & 39.31 & 25.06 & 42.17 & 39.14 & 33.37 & 24.07 & 30.99 & 21.97 & 29.17 & 39.43 & 49.93 & 39.14 & 38.79 & 35.09 & 29.63 & 34.18 \\
\rowcolor{green!20} FSF-Net & Next-1 & 48.15 & 42.38 & 46.61 & 34.29 & 51.30 & 47.66 & 44.45 & 34.39 & 32.99 & 31.49 & 38.90 & 38.90 & 53.71 & 44.20 & 44.61 & 41.68 & 40.66 & 45.42 \\
    \hline
Copy-Paste & Next-2 & 24.56 & 15.01 & 16.75 & 4.91 & 20.36 & 14.83 & 11.03 & 5.03 & 3.29 & 6.45 & 13.65 & 12.72 & 39.64 & 24.92 & 23.67 & 21.80 & 13.70 & 14.25\\
OccWorld \cite{occ-world} & Next-2 & 29.74 & 20.06 & 25.28 & 25.28 & 20.97 & 24.01 & 18.87 & 8.22 & 14.49 & 9.56 & 13.21 & 21.74 & 37.55 & 26.80 & 27.67 & 22.03 & 19.92 & 24.49 \\
BEV-Flow & Next-2 & 26.08 & 16.53 & 13.00 & 7.35 & 16.61 & 12.83 & 19.27 & 6.96 & 7.07 & 5.95 & 14.59 & 15.89 & 40.30 & 23.47 & 27.12 & 24.66 & 12.53 & 21.20 \\
Base-Net & Next-2 & 27.37 & 21.28 & 26.14 & 11.75 & 27.36 & 24.77 & 21.36 & 11.15 & 14.85 & 10.14 & 16.33 & 24.21 & 40.61 & 29.54 & 27.06 & 22.91 & 17.04 & 19.24 \\
\rowcolor{green!20} FSF-Net & Next-2 & 36.01 & 28.60 & 31.73 & 19.21 & 36.75 & 32.87 & 31.56 & 18.34 & 16.88 & 16.93 & 24.12 & 34.02 & 43.71 & 32.76 & 33.11 & 29.83 & 26.56 & 32.75 \\
   \hline
Copy-Paste & Next-3 & 21.71 & 12.31 & 13.80 & 4.10 & 15.44 & 12.00 & 8.30 & 3.68 & 2.53 & 5.69 & 10.91 & 9.90 & 34.90 & 21.20 & 19.79 & 17.65 & 11.50 & 12.09 \\
OccWorld \cite{occ-world} & Next-3 & 24.64 & 14.22 & 18.76 & 4.94 & 11.93 & 16.90 & 12.60 & 3.81 & 7.84 & 5.70 & 8.08 & 14.19 & 31.94 & 21.23 & 22.15 & 16.44 & 15.28 & 18.89 \\
BEV-Flow & Next-3 & 19.00 & 9.67 & 5.69 & 2.52 & 8.72 & 7.23 & 10.01 & 2.90 & 3.37 & 2.23 & 8.73 & 8.43 & 31.53 & 13.43 & 17.57 & 15.91 & 6.91 & 13.00 \\
Base-Net & Next-1 & 21.30 & 15.12 & 19.96 & 6.08 & 18.89 & 17.35 & 15.20 & 6.01 & 8.34 & 6.06 & 10.17 & 16.51 & 35.27 & 24.78 & 20.48 & 16.35 & 11.83 & 11.78 \\
\rowcolor{green!20} FSF-Net & Next-3 & 29,48 & 21.34 & 23.97 & 11.49 & 27.62 & 24.16 & 23.95 & 10.30 & 10.08 & 10.19 & 16.72 & 24.93 & 38.48 & 26.85 & 27.00 & 23.60 & 19.74 & 25.08 \\
   \hline
Copy-Paste & Next-4 & 19.85 & 10.66 & 11.76 & 3.50 & 12.25 & 10.36 & 6.82 & 3.04 & 2.20 & 5.09 & 9.63 & 8.67 & 31.82 & 18.44 & 17.37 & 15.11 & 10.11 & 10.72 \\
OccWorld \cite{occ-world} & Next-4 & 21.35 & 10.84 & 14.55 & 2.46 & 7.01 & 12.72 & 8.82 & 1.84 & 4.74 & 3.91 & 5.46 & 9.83 & 28.08 & 17.98 & 18.40 & 12.92 & 12.46 & 15.06 \\
BEV-Flow & Next-4 & 14.93 & 6.09 & 2.97 & 1.22 & 4.45 & 4.20 & 5.00 & 1.55 & 1.79 & 1.06 & 5.66 & 4.85 & 25.01 & 7.47 & 11.25 & 10.58 & 4.57 & 8.46 \\
Base-Net & Next-1 & 17.47 & 11.427 & 16.16 & 3.26 & 13.39 & 12.96 & 10.94 & 3.59 & 5.25 & 4.46 & 6.52 & 11.89 & 31.25 & 21.33 & 15.81 & 12.02 & 8.88 & 7.41 \\
\rowcolor{green!20} FSF-Net & Next-4 & 25.17 & 17.03 & 19.46 & 7.72 & 21.32 & 18.63 & 19.41 & 6.42 & 6.56 & 7.04 & 12.69 & 19.45 & 34.93 & 22.79 & 22.95 & 19.46 & 15.87 & 19.98 \\
    \hline
\end{tabular}}}
\label{tab:exp_1}
\end{table*}
\newpage

\begin{table*}[ht]
\centering
\caption{Performance of BEV occupancy forecasting in Occ3D validation dataset. Term Next-$A$ means forecasting the next $A$-th frame.}
\resizebox{1.0\linewidth}{!}{
\rotatebox{0}{
\begin{tabular}{c| c | c c | c c c c c c c c c c c c c c c c}
    \hline
    \textbf{Method}
    & {Frame} 
    & IoU& mIoU
    & \rotatebox{90}{\textcolor{nbarrier}{$\blacksquare$} barrier}
    & \rotatebox{90}{\textcolor{nbicycle}{$\blacksquare$} bicycle}
    & \rotatebox{90}{\textcolor{nbus}{$\blacksquare$} bus}
    & \rotatebox{90}{\textcolor{ncar}{$\blacksquare$} car}
    & \rotatebox{90}{\textcolor{nconstruct}{$\blacksquare$} const. veh.}
    & \rotatebox{90}{\textcolor{nmotor}{$\blacksquare$} motorcycle}
    & \rotatebox{90}{\textcolor{npedestrian}{$\blacksquare$} pedestrian}
    & \rotatebox{90}{\textcolor{ntraffic}{$\blacksquare$} traffic cone}
    & \rotatebox{90}{\textcolor{ntrailer}{$\blacksquare$} trailer}
    & \rotatebox{90}{\textcolor{ntruck}{$\blacksquare$} truck}
    & \rotatebox{90}{\textcolor{ndriveable}{$\blacksquare$} drive. suf.}
    & \rotatebox{90}{\textcolor{nother}{$\blacksquare$} other flat}
    & \rotatebox{90}{\textcolor{nsidewalk}{$\blacksquare$} sidewalk}
    & \rotatebox{90}{\textcolor{nterrain}{$\blacksquare$} terrain}
    & \rotatebox{90}{\textcolor{nmanmade}{$\blacksquare$} manmade}
    & \rotatebox{90}{\textcolor{nvegetation}{$\blacksquare$} vegetation} \\
    \hline
Copy-Paste & Next-1 & 68.43 & 26.70 & 24.07 & 10.03 & 33.29 & 22.01 & 24.20 & 5.81 & 9.09 & 11.42 & 26.01 & 25.56 & 67.70 & 43.13 & 38.63 & 40.35 & 26.71 & 36.50 \\
OccWorld \cite{occ-world} & Next-1 & 72.69 & 35.45 & 31.28 & 16.37 & 36.93 & 36.76 & 40.34 & 15.30 & 26.00 & 17.05 & 28.45 & 38.71 & 74.31 & 54.48 & 52.27 & 49.89 & 35.13 & 48.41 \\
BEV-Flow & Next-1 & 76.52 & 41.87 & 35.29 & 18.54 & 44.84 & 33.23 & 54.41 & 17.68 & 21.12 & 22.09 & 40.34 & 44.54 & 77.72 & 57.71 & 58.12 & 58.44 & 43.87 & 61.99 \\
Base-Net & Next-1 & 68.03 & 37.32 & 33.73 & 20.80 & 41.67 & 37.45 & 42.82 & 20.06 & 26.88 & 18.48 & 31.21 & 41.48 & 75.43 & 56.41 & 53.73 & 50.63 & 30.38 & 41.00 \\
\rowcolor{green!20} FSF-Net & Next-1 & 76.76 & 47.55 & 41.61 & 27.92 & 53.29 & 47.38 & 56.72 & 30.87 & 30.20 & 26.05 & 44.76 & 52.36 & 79.57 & 64.08 & 62.70 & 60.78 & 48.00 & 60.23 \\
    \hline
Copy-Paste & Next-2 & 64.60 & 19.80 & 18.22 & 6.86 & 21.75 & 15.64 & 15.39 & 3.31 & 4.44 & 9.36 & 18.16 & 16.00 & 59.50 & 33.50 & 30.17 & 32.17 & 20.66 & 28.73 \\
OccWorld \cite{occ-world} & Next-2 & 72.68 & 35.45 & 21.84 & 7.33 & 20.13 & 24.75 & 27.20 & 5.87 & 13.36 & 8.97 & 16.24 & 25.10 & 67.50 & 42.38 & 43.16 & 40.03 & 26.22 & 40.17 \\
BEV-Flow & Next-2 & 67.80 & 25.02 & 15.33 & 6.36 & 25.01 & 17.52 & 32.02 & 5.54 & 7.65 & 7.12 & 22.59 & 24.67 & 67.11 & 37.29 & 39.35 & 41.81 & 25.03 & 44.02 \\
Base-Net & Next-2 & 56.82 & 24.22 & 21.18 & 9.93 & 26.96 & 23.54 & 27.44 & 8.93 & 13.44 & 9.30 & 17.43 & 25.91 & 66.81 & 45.24 & 40.04 & 35.77 & 16.27 & 21.86 \\
\rowcolor{green!20} FSF-Net & Next-2  & 68.76 & 34.33 & 28.33 & 16.03 & 39.36 & 33.43 & 42.83 & 16.50 & 16.29 & 14.60 & 29.56 & 37.94 & 72.38 & 52.07 & 50.72 & 48.81 & 32.24 & 44.98 \\
   \hline
Copy-Paste & Next-3 & 62.24 & 16.86 & 15.84 & 6.06 & 16.37 & 13.10 & 11.57 & 2.02 & 3.62 & 8.59 & 15.02 & 12.74 & 54.82 & 28.93 & 26.06 & 27.94 & 17.77 & 24.93 \\
OccWorld \cite{occ-world} & Next-3 & 66.93 & 19.57 & 16.49 & 3.55 & 12.01 & 17.85 & 18.48 & 2.74 & 7.51 & 5.77 & 10.59 & 17.40 & 61.60 & 34.60 & 36.40 & 33.04 & 21.08 & 33.52 \\
BEV-Flow & Next-3 & 62.67 & 16.58 & 8.06 & 2.90 & 13.90 & 10.41 & 18.11 & 1.78 & 4.02 & 3.04 & 14.54 & 58.04 & 58.04 & 23.24 & 26.94 & 31.12 & 16.84 & 31.90 \\
Base-Net & Next-3 & 49.80 & 17.80 & 15.43 & 5.27 & 18.25 & 16.48 & 18.97 & 4.77 & 7.91 & 6.08 & 10.77 & 17.85 & 60.32 & 38.54 & 31.30 & 26.97 & 10.75 & 12.79 \\
\rowcolor{green!20} FSF-Net & Next-3 & 63.75 & 26.67 & 21.13 & 10.09 & 30.09 & 24.96 & 33.48 & 9.34 & 9.85 & 9.13 & 21.28 & 28.52 & 66.92 & 43.52 & 43.02 & 41.18 & 24.28 & 35.53 \\
   \hline
Copy-Paste & Next-4 & 60.44 & 15.07 & 14.04 & 5.56 & 13.00 & 11.62 & 9.63 & 1.55 & 3.31 & 7.96 & 13.13 & 11.37 & 51.63 & 25.42 & 23.62 & 25.08 & 15.08 & 22.65 \\
OccWorld \cite{occ-world} & Next-4 & 64.93 & 15.94 & 12.96 & 1.65 & 7.35 & 13.73 & 13.26 & 1.31 & 4.60 & 4.16 & 7.49 & 12.62 & 56.68 & 29.78 & 31.31 & 27.85 & 17.66 & 27.52 \\
BEV-Flow & Next-4 & 58.80 & 11.57 & 4.99 & 1.72 & 7.07 & 6.31 & 9.80 & 0.95 & 2.33 & 1.57 & 9.59 & 8.59 & 49.97 & 13.67 & 18.45 & 23.84 & 12.55 & 23.63 \\
Base-Net & Next-4& 44.39 & 13.85 & 11.79 & 3.21 & 12.75 & 12.37 & 12.98 & 2.76 & 5.21 & 4.54 & 7.09 & 12.97 & 54.67 & 33.67 & 24.78 & 20.61 & 7.82 & 7.84 \\
\rowcolor{green!20} FSF-Net & Next-4& 60.08 & 22.04 & 17.22 & 7.47 & 23.62 & 19.39 & 26.89 & 6.08 & 6.51 & 6.39 & 16.15 & 22.37 & 62.44 & 37.34 & 37.40 & 35.81 & 19.58 & 29.06 \\
    \hline
\end{tabular}}}
\label{tab:exp_2}
\end{table*}

\begin{table*}[ht]
\centering
\caption{Performance of 4D occupancy forecasting in Occ3D testing dataset. Term Next-$A$ means forecasting the next $A$-th frame.}
\resizebox{1.0\linewidth}{!}{
\rotatebox{0}{
\begin{tabular}{c| c | c c | c c c c c c c c c c c c c c c c}
    \hline
    \textbf{Method}
    & {Frame} 
    & IoU& mIoU
    & \rotatebox{90}{\textcolor{nbarrier}{$\blacksquare$} barrier}
    & \rotatebox{90}{\textcolor{nbicycle}{$\blacksquare$} bicycle}
    & \rotatebox{90}{\textcolor{nbus}{$\blacksquare$} bus}
    & \rotatebox{90}{\textcolor{ncar}{$\blacksquare$} car}
    & \rotatebox{90}{\textcolor{nconstruct}{$\blacksquare$} const. veh.}
    & \rotatebox{90}{\textcolor{nmotor}{$\blacksquare$} motorcycle}
    & \rotatebox{90}{\textcolor{npedestrian}{$\blacksquare$} pedestrian}
    & \rotatebox{90}{\textcolor{ntraffic}{$\blacksquare$} traffic cone}
    & \rotatebox{90}{\textcolor{ntrailer}{$\blacksquare$} trailer}
    & \rotatebox{90}{\textcolor{ntruck}{$\blacksquare$} truck}
    & \rotatebox{90}{\textcolor{ndriveable}{$\blacksquare$} drive. suf.}
    & \rotatebox{90}{\textcolor{nother}{$\blacksquare$} other flat}
    & \rotatebox{90}{\textcolor{nsidewalk}{$\blacksquare$} sidewalk}
    & \rotatebox{90}{\textcolor{nterrain}{$\blacksquare$} terrain}
    & \rotatebox{90}{\textcolor{nmanmade}{$\blacksquare$} manmade}
    & \rotatebox{90}{\textcolor{nvegetation}{$\blacksquare$} vegetation} \\
    \hline
Copy-Paste & Next-1 & 30.16 & 21.20 & 23.34 & 8.11 & 31.62 & 20.98 & 17.67 & 7.25 & 7.11 & 8.54 & 19.12 & 20.94 & 48.96 & 32.80 & 32.30 & 30.47 & 18.28 & 18.54 \\
OccWorld & Next-1 & 38.79 & 32.17 & 38.51 & 21.03 & 41.45 & 37.66 & 31.81 & 20.70 & 30.37 & 21.22 & 25.04 & 37.62 & 47.85 & 36.91 & 37.24 & 32.95 & 29.41 & 32.28 \\
BEV-Flow & Next-1 & 41.51 & 32.98 & 35.81 & 22.12 & 34.87 & 27.66 & 38.89 & 18.64 & 21.25 & 22.15 & 30.45 & 34.15 & 53.77 & 40.88 & 43.13 & 40.25 & 29.60 & 37.78 \\
Base-Net & Next-1 & 40.07 & 34.66 & 40.04 & 25.50 & 41.03 & 39.63 & 34.83 & 24.39 & 31.71 & 22.59 & 30.80 & 39.35 & 50.36 & 41.86 & 39.34 & 36.02 & 29.99 & 34.11 \\
\rowcolor{green!20} FSF-Net & Next-1 & 49.95 & 44.05 & 47.88 & 35.70 & 53.19 & 48.70 & 43.81 & 36.61 & 35.36 & 31.53 & 39.54 & 50.85 & 56.79 & 46.87 & 46.99 & 45.20 & 41.55 & 46.30 \\
    \hline
    \textbf{Method}
    & {Frame} 
    & BEVIoU& BEVmIoU
    & \rotatebox{90}{\textcolor{nbarrier}{$\blacksquare$} barrier}
    & \rotatebox{90}{\textcolor{nbicycle}{$\blacksquare$} bicycle}
    & \rotatebox{90}{\textcolor{nbus}{$\blacksquare$} bus}
    & \rotatebox{90}{\textcolor{ncar}{$\blacksquare$} car}
    & \rotatebox{90}{\textcolor{nconstruct}{$\blacksquare$} const. veh.}
    & \rotatebox{90}{\textcolor{nmotor}{$\blacksquare$} motorcycle}
    & \rotatebox{90}{\textcolor{npedestrian}{$\blacksquare$} pedestrian}
    & \rotatebox{90}{\textcolor{ntraffic}{$\blacksquare$} traffic cone}
    & \rotatebox{90}{\textcolor{ntrailer}{$\blacksquare$} trailer}
    & \rotatebox{90}{\textcolor{ntruck}{$\blacksquare$} truck}
    & \rotatebox{90}{\textcolor{ndriveable}{$\blacksquare$} drive. suf.}
    & \rotatebox{90}{\textcolor{nother}{$\blacksquare$} other flat}
    & \rotatebox{90}{\textcolor{nsidewalk}{$\blacksquare$} sidewalk}
    & \rotatebox{90}{\textcolor{nterrain}{$\blacksquare$} terrain}
    & \rotatebox{90}{\textcolor{nmanmade}{$\blacksquare$} manmade}
    & \rotatebox{90}{\textcolor{nvegetation}{$\blacksquare$} vegetation} \\
    \hline
Copy-Paste & Next-1 & 68.43 & 26.70 & 24.07 & 10.03 & 33.29 & 22.01 & 24.20 & 5.81 & 9.09 & 11.42 & 26.01 & 25.56 & 67.70 & 43.13 & 38.63 & 40.35 & 26.71 & 36.50 \\
OccWorld \cite{occ-world} & Next-1 & 72.65 & 35.80 & 33.18 & 17.35 & 37.66 & 37.25 & 38.46 & 15.94 & 27.06 & 18.50 & 27.07 & 39.59 & 74.33 & 55.00 &  52.16 & 50.54 & 35.62 & 40.38 \\
BEV-Flow & Next-1 & 76.52 & 41.87 & 35.29 & 18.54 & 44.84 & 33.23 & 54.41 & 17.68 & 21.12 & 22.09 & 40.34 & 44.54 & 77.72 & 57.71 & 58.12 & 58.44 & 43.87 & 61.99 \\
Base-Net & Next-1 & 68.26 & 37.82 & 35.46 & 22.18 & 40.84 & 37.60 & 44.29 & 18.56 & 26.42 & 18.79 & 31.15 & 41.74 & 75.91 & 56.81 & 54.40 & 50.80 & 30.24 & 41.19 \\
\rowcolor{green!20} FSF-Net & Next-1 & 78.18 & 48.21 & 44.50 & 30.95 & 51.74 & 47.94 & 56.32 & 31.35 & 31.87 & 26.35 & 43.97 & 51.88 & 80.74 & 65.52 & 63.69 & 62.92 & 49.58 & 62.04 \\
    \hline
\end{tabular}}}
\label{tab:exp_3}
\end{table*}

\begin{figure*}[!ht]
	\centering
		\includegraphics[width=1.0\linewidth]{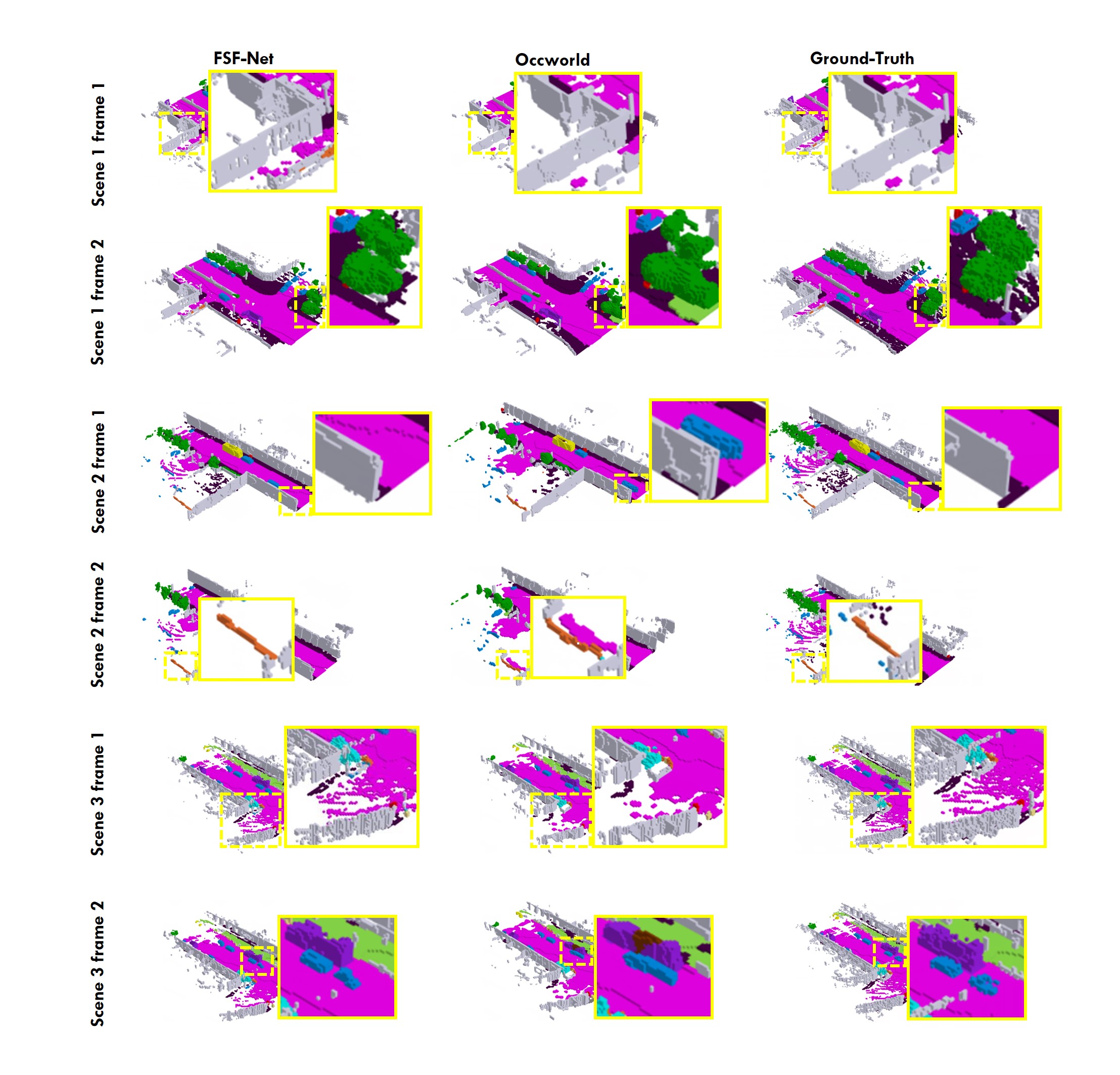}	
		\caption{Visualization of the prediction results of the first two frames for three scenarios by OccWorld and FSF-Net.}
\label{fig:2_frame}
\end{figure*}

\begin{figure*}[ht]
	\centering
		\includegraphics[width=1.0\linewidth]{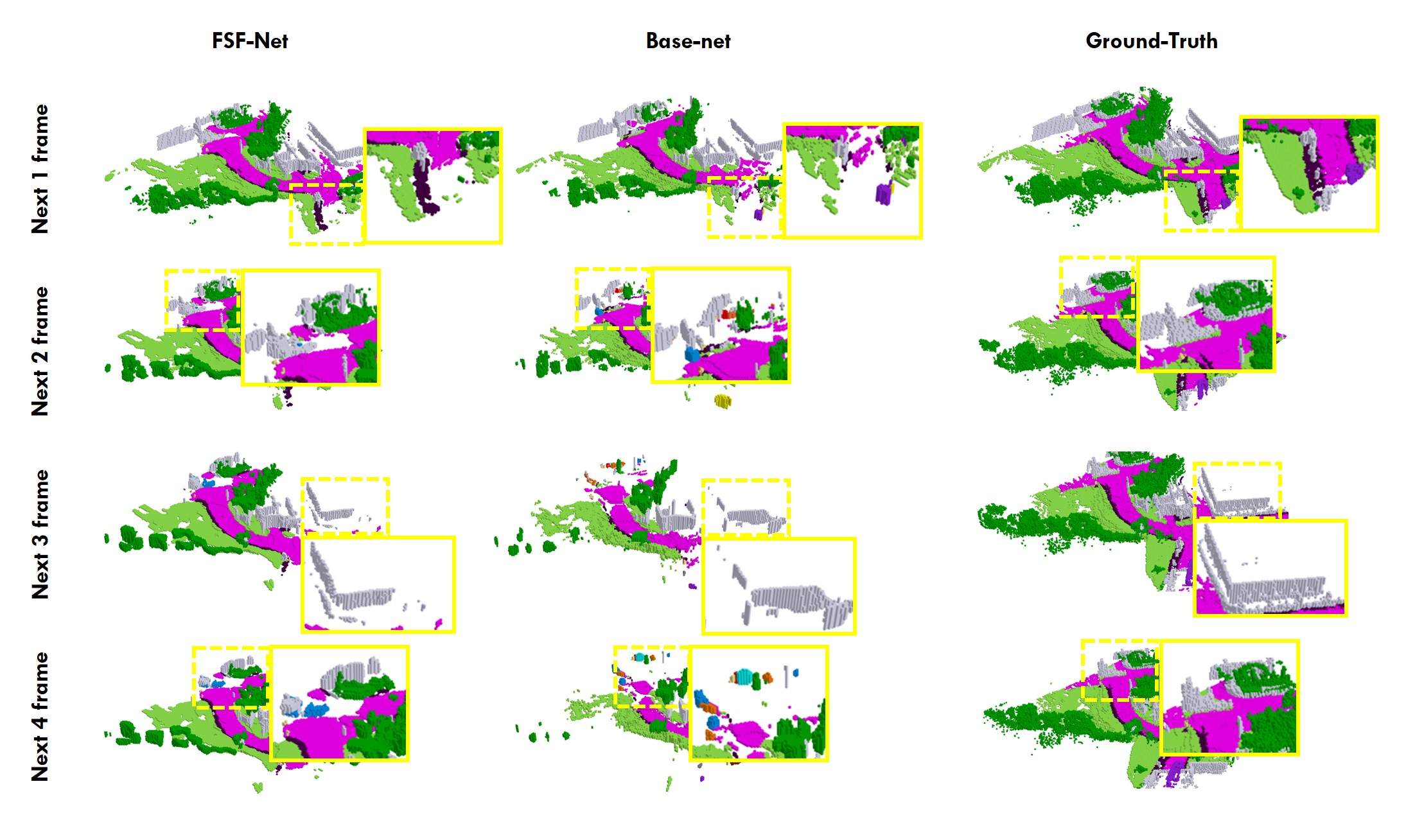}	
		\caption{Visualization of the four-frame prediction results for the same scenario by Base-NAT and FSF-Net.}
\label{fig:4_frame}
\end{figure*}

\begin{table*}[ht]
\centering
\caption{Performance of 4D occupancy forecasting in Occ3D validation dataset when only 20\% training dataset is used for model training. Term Next-$A$ means forecasting the next $A$-th frame.}
\resizebox{1.0\linewidth}{!}{
\rotatebox{0}{
\begin{tabular}{c| c | c c | c c c c c c c c c c c c c c c c}
    \hline
    \textbf{Method}
    & {Frame} 
    & IoU& mIoU
    & \rotatebox{90}{\textcolor{nbarrier}{$\blacksquare$} barrier}
    & \rotatebox{90}{\textcolor{nbicycle}{$\blacksquare$} bicycle}
    & \rotatebox{90}{\textcolor{nbus}{$\blacksquare$} bus}
    & \rotatebox{90}{\textcolor{ncar}{$\blacksquare$} car}
    & \rotatebox{90}{\textcolor{nconstruct}{$\blacksquare$} const. veh.}
    & \rotatebox{90}{\textcolor{nmotor}{$\blacksquare$} motorcycle}
    & \rotatebox{90}{\textcolor{npedestrian}{$\blacksquare$} pedestrian}
    & \rotatebox{90}{\textcolor{ntraffic}{$\blacksquare$} traffic cone}
    & \rotatebox{90}{\textcolor{ntrailer}{$\blacksquare$} trailer}
    & \rotatebox{90}{\textcolor{ntruck}{$\blacksquare$} truck}
    & \rotatebox{90}{\textcolor{ndriveable}{$\blacksquare$} drive. suf.}
    & \rotatebox{90}{\textcolor{nother}{$\blacksquare$} other flat}
    & \rotatebox{90}{\textcolor{nsidewalk}{$\blacksquare$} sidewalk}
    & \rotatebox{90}{\textcolor{nterrain}{$\blacksquare$} terrain}
    & \rotatebox{90}{\textcolor{nmanmade}{$\blacksquare$} manmade}
    & \rotatebox{90}{\textcolor{nvegetation}{$\blacksquare$} vegetation} \\
    \hline
Copy-Paste & Next-1 & 30.16 & 21.20 & 23.34 & 8.11 & 31.62 & 20.98 & 17.67 & 7.25 & 7.11 & 8.54 & 19.12 & 20.94 & 48.96 & 32.80 & 32.30 & 30.47 & 18.28 & 18.54 \\
OccWorld \cite{occ-world} & Next-1 & 34.83 & 21.91 & 17.24 & 5.05 & 26.23 & 27.01 & 21.10 & 12.40 & 13.81 & 8.33 & 14.89 & 27.37 & 45.60 & 30.74 & 31.69 & 30.39 & 24.11 & 29.11 \\
BEV-Flow & Next-1  & 41.51 & 32.98 & 35.81 & 22.12 & 34.87 & 27.66 & 38.89 & 18.64 & 21.25 & 22.15 & 30.45 & 34.15 & 53.77 & 40.88 & 43.13 & 40.25 & 29.60 & 37.78 \\
Base-Net & Next-1  & 36.00 & 23.76 & 17.60 & 6.76 & 28.64 & 28.81 & 26.15 & 14.57 & 14.57 & 9.35 & 16.69 & 29.82 & 47.24 & 32.05 & 33.62 & 32.36 & 25.28 & 31.12 \\
\rowcolor{green!20} FSF-Net & Next-1 & 40.50 & 30.19 & 27.16 & 13.90 & 36.90 & 35.03 & 30.97 & 20.41 & 23.41 & 18.58 & 20.61 & 36.38 & 50.16 & 36.53 & 40.23 & 37.03 & 28.94 & 36.41 \\
    \hline
Copy-Paste & Next-2  & 24.56 & 15.01 & 16.75 & 4.91 & 20.36 & 14.83 & 11.03 & 5.03 & 3.29 & 6.45 & 13.65 & 12.72 & 39.64 & 24.92 & 23.67 & 21.80 & 13.70 & 14.25\\
OccWorld \cite{occ-world} & Next-2 & 26.40 & 13.74 & 10.22 & 1.68 & 14.25 & 15.44 & 13.38 & 4.05 & 5.12 & 2.81 & 7.38 & 15.48 & 37.29 & 21.36 & 22.90 & 21.53 & 15.72 & 20.78 \\
BEV-Flow & Next-2  & 26.08 & 16.53 & 13.00 & 7.35 & 16.61 & 12.83 & 19.27 & 6.96 & 7.07 & 5.95 & 14.59 & 15.89 & 40.30 & 23.47 & 27.12 & 24.66 & 12.53 & 21.20 \\
Base-Net & Next-2  & 23.31 & 12.93 & 8.98 & 1.62 & 13.97 & 15.50 & 14.31 & 4.66 & 5.03 & 2.75 & 5.84 & 15.52 & 37.18 & 22.10 & 20.50 & 19.50 & 13.89 & 14.93 \\
\rowcolor{green!20} FSF-Net & Next-2 & 29.14 & 17.95 & 13.99 & 5.69 & 21.89 & 21.33 & 19.66 & 9.26 & 9.59 & 6.89 & 8.59 & 21.05 & 40.43 & 23.97 & 29.30 & 26.16 & 15.69 & 24.27 \\
    \hline
Copy-Paste & Next-3   & 21.71 & 12.31 & 13.80 & 4.10 & 15.44 & 12.00 & 8.30 & 3.68 & 2.53 & 5.69 & 10.91 & 9.90 & 34.90 & 21.20 & 19.79 & 17.65 & 11.50 & 12.09 \\
OccWorld \cite{occ-world} & Next-3 & 21.84 & 9.86 & 6.70 & 0.65 & 8.64 & 10.24 & 8.97 & 1.62 & 2.54 & 1.55 & 4.39 & 9.65 & 32.42 & 16.31 & 17.69 & 16.45 & 11.79 & 15.62 \\
BEV-Flow & Next-3  & 19.00 & 9.67 & 5.69 & 2.52 & 8.72 & 7.23 & 10.01 & 2.90 & 3.37 & 2.23 & 8.73 & 8.43 & 31.53 & 13.43 & 17.57 & 15.91 & 6.91 & 13.00 \\
Base-Net & Next-3  & 17.97 & 8.60 & 5.71 & 0.62 & 7.91 & 9.98 & 8.85 & 1.75 & 2.49 & 1.29 & 2.61 & 9.24 & 32.27 & 16.74 & 13.96 & 13.15 & 9.77 & 8.16 \\
\rowcolor{green!20} FSF-Net & Next-3 &23.70 & 12.58 & 8.94 & 2.62 & 13.93 & 14.75 & 13.49 & 4.89 & 5.23 & 3.14 & 4.15 & 13.36 & 35.39 & 17.87 & 23.40 & 20.51 & 10.49 & 17.76 \\
    \hline
Copy-Paste & Next-4   & 19.85 & 10.66 & 11.76 & 3.50 & 12.25 & 10.36 & 6.82 & 3.04 & 2.20 & 5.09 & 9.63 & 8.67 & 31.82 & 18.44 & 17.37 & 15.11 & 10.11 & 10.72 \\
OccWorld \cite{occ-world} & Next-4 & 19.11 & 7.66 & 4.73 & 0.28 & 5.42 & 7.63 & 5.82 & 0.76 & 1.56 & 0.95 & 3.09 & 6.37 & 29.16 & 13.06 & 14.47 & 13.30 & 9.53 & 12.50 \\
BEV-Flow & Next-4  & 14.93 & 6.09 & 2.97 & 1.22 & 4.45 & 4.20 & 5.00 & 1.55 & 1.79 & 1.06 & 5.66 & 4.85 & 25.01 & 7.47 & 11.25 & 10.58 & 4.57 & 8.46 \\
Base-Net & Next-4  & 15.09 & 6.33 & 3.92 & 0.30 & 4.78 & 7.23 & 5.66 & 0.92 & 1.40 & 0.73 & 1.28 & 5.96 & 29.41 & 13.31 & 9.93 & 9.39 & 7.67 & 4.73 \\
\rowcolor{green!20} FSF-Net & Next-4 & 20.32 & 9.57 & 6.13 & 1.35 & 9.34 & 10.89 & 9.66 & 2.89 & 3.30 & 1.75 & 2.15 & 9.28 & 31.92 & 13.94 & 19.39 & 16.99 & 7.80 & 13.52 \\
    \hline
\end{tabular}}}
\label{tab:exp_4}
\end{table*}

\begin{table*}[ht]
\centering
\caption{Performance of BEV occupancy forecasting in Occ3D validation dataset when only 20\% training dataset is used for model training. Term Next-$A$ means forecasting the next $A$-th frame.}
\resizebox{1.0\linewidth}{!}{
\rotatebox{0}{
\begin{tabular}{c| c | c c | c c c c c c c c c c c c c c c c}
    \hline
    \textbf{Method}
    & {Frame} 
    & IoU& mIoU
    & \rotatebox{90}{\textcolor{nbarrier}{$\blacksquare$} barrier}
    & \rotatebox{90}{\textcolor{nbicycle}{$\blacksquare$} bicycle}
    & \rotatebox{90}{\textcolor{nbus}{$\blacksquare$} bus}
    & \rotatebox{90}{\textcolor{ncar}{$\blacksquare$} car}
    & \rotatebox{90}{\textcolor{nconstruct}{$\blacksquare$} const. veh.}
    & \rotatebox{90}{\textcolor{nmotor}{$\blacksquare$} motorcycle}
    & \rotatebox{90}{\textcolor{npedestrian}{$\blacksquare$} pedestrian}
    & \rotatebox{90}{\textcolor{ntraffic}{$\blacksquare$} traffic cone}
    & \rotatebox{90}{\textcolor{ntrailer}{$\blacksquare$} trailer}
    & \rotatebox{90}{\textcolor{ntruck}{$\blacksquare$} truck}
    & \rotatebox{90}{\textcolor{ndriveable}{$\blacksquare$} drive. suf.}
    & \rotatebox{90}{\textcolor{nother}{$\blacksquare$} other flat}
    & \rotatebox{90}{\textcolor{nsidewalk}{$\blacksquare$} sidewalk}
    & \rotatebox{90}{\textcolor{nterrain}{$\blacksquare$} terrain}
    & \rotatebox{90}{\textcolor{nmanmade}{$\blacksquare$} manmade}
    & \rotatebox{90}{\textcolor{nvegetation}{$\blacksquare$} vegetation} \\
    \hline
Copy-Paste & Next-1  & 68.43 & 26.70 & 24.07 & 10.03 & 33.29 & 22.01 & 24.20 & 5.81 & 9.09 & 11.42 & 26.01 & 25.56 & 67.70 & 43.13 & 38.63 & 40.35 & 26.71 & 36.50 \\
OccWorld \cite{occ-world} & Next-1 & 68.00 & 27.26 & 16.32 & 3.09 & 28.20 & 27.80 & 31.32 & 11.45 & 13.11 & 7.33 & 16.94 & 31.13 & 71.85 & 40.95 & 44.90 & 46.88 & 29.36 & 42.87 \\
BEV-Flow & Next-1  & 76.52 & 41.87 & 35.29 & 18.54 & 44.84 & 33.23 & 54.41 & 17.68 & 21.12 & 22.09 & 40.34 & 44.54 & 77.72 & 57.71 & 58.12 & 58.44 & 43.87 & 61.99 \\
Base-Net & Next-1  & 64.67 & 27.55 & 17.42 & 5.79 & 27.80 & 27.87 & 33.40 & 13.95 & 13.71 & 8.17 & 16.22 & 31.85 & 72.73 & 44.08 & 46.18 & 47.47 & 25.69 & 36.18 \\
\rowcolor{green!20} FSF-Net & Next-1 & 70.46 & 34.47 & 24.31 & 9.83 & 36.40 & 34.88 & 42.61 & 18.88 & 21.40 & 16.09 & 20.53 & 38.86 & 75.99 & 49.30 & 54.26 & 54.00 & 30.70 & 48.36 \\
    \hline
Copy-Paste & Next-2  & 64.60 & 19.80 & 18.22 & 6.86 & 21.75 & 15.64 & 15.39 & 3.31 & 4.44 & 9.36 & 18.16 & 16.00 & 59.50 & 33.50 & 30.17 & 32.17 & 20.66 & 28.73 \\
OccWorld \cite{occ-world} & Next-2 & 63.92 & 13.74 & 9.85 & 1.01 & 16.69 & 17.05 & 20.84 & 4.00 & 4.78 & 2.31 & 8.96 & 19.44 & 66.00 & 30.98 & 35.62 & 37.98 & 21.41 & 34.28 \\
BEV-Flow & Next-2  & 67.80 & 25.02 & 15.33 & 6.36 & 25.01 & 17.52 & 32.02 & 5.54 & 7.65 & 7.12 & 22.59 & 24.67 & 67.11 & 37.29 & 39.35 & 41.81 & 25.03 & 44.02 \\
Base-Net & Next-2  & 52.65 & 16.41 & 8.98 & 1.55 & 13.83 & 14.83 & 19.01 & 4.62 & 4.49 & 2.46 & 5.29 & 16.69 & 64.12 & 32.60 & 29.90 & 30.91 & 13.06 & 16.73 \\
\rowcolor{green!20} FSF-Net & Next-2 & 63.25 & 22.90 & 7.89 & 1.52 & 14.84 & 15.24 & 19.79 & 4.44 & 4.85 & 3.11 & 4.50 & 16.24 & 63.13 & 26.06 & 35.06 & 35.09 & 11.45 & 26.45 \\
    \hline
Copy-Paste & Next-3  & 62.24 & 16.86 & 15.84 & 6.06 & 16.37 & 13.10 & 11.57 & 2.02 & 3.62 & 8.59 & 15.02 & 12.74 & 54.82 & 28.93 & 26.06 & 27.94 & 17.77 & 24.93 \\
OccWorld \cite{occ-world} & Next-3 & 61.38 & 15.17 & 6.26 & 0.56 & 10.06 & 11.79 & 14.36 & 1.59 & 2.24 & 1.33 & 5.64 & 12.35 & 61.22 & 24.45 & 29.13 & 31.88 & 13.19 & 27.91 \\
BEV-Flow & Next-3  & 62.67 & 16.58 & 8.06 & 2.90 & 13.90 & 10.41 & 18.11 & 1.78 & 4.02 & 3.04 & 14.54 & 58.04 & 58.04 & 23.24 & 26.94 & 31.12 & 16.84 & 31.90 \\
Base-Net & Next-3  & 46.18 & 11.61 & 5.88 & 0.61 & 7.99 & 9.56 & 11.50 & 1.91 & 2.09 & 1.16 & 2.33 & 9.85 & 58.69 & 25.58 & 20.84 & 21.66 & 8.82 & 8.91 \\
\rowcolor{green!20} FSF-Net & Next-3 & 58.94 & 17.18 & 7.89 & 1.52 & 14.84 & 15.24 & 19.79 & 4.44 & 4.85 & 3.11 & 4.50 & 16.24 & 63.13 & 26.06 & 35.06 & 35.09 & 11.45 & 26.45 \\
    \hline
Copy-Paste & Next-4  & 60.44 & 15.07 & 14.04 & 5.56 & 13.00 & 11.62 & 9.63 & 1.55 & 3.31 & 7.96 & 13.13 & 11.37 & 51.63 & 25.42 & 23.62 & 25.08 & 15.08 & 22.65 \\
OccWorld \cite{occ-world} & Next-4 & 19.11 & 7.66 & 4.47 & 0.31 & 6.30 & 8.97 & 9.41 & 0.68 & 1.34 & 0.74 & 4.12 & 8.33 & 57.40 & 19.79 & 24.73 & 27.65 & 14.58 & 23.70 \\
BEV-Flow & Next-4  & 58.80 & 11.57 & 4.99 & 1.72 & 7.07 & 6.31 & 9.80 & 0.95 & 2.33 & 1.57 & 9.59 & 8.59 & 49.97 & 13.67 & 18.45 & 23.84 & 12.55 & 23.63 \\
Base-Net & Next-4  & 41.79 & 8.92 & 4.13 & 0.35 & 4.97 & 6.85 & 7.19 & 1.10 & 1.16 & 0.75 & 1.28 & 6.24 & 54.61 & 20.68 & 14.92 & 15.59 & 6.75 & 5.12 \\
\rowcolor{green!20} FSF-Net & Next-4 & 55.62 & 13.70 & 5.61 & 1.01 & 10.34 & 11.33 & 14.45 & 2.47 & 3.07 & 1.66 & 2.12 & 11.60 & 58.46 & 20.47 & 29.94 & 30.26 & 8.41 & 20.78 \\
    \hline
\end{tabular}}}
\label{tab:exp_5}
\end{table*}

\textbf{Results on Evaluation of 4D Occupancy Forecasting.} We first trained our model using the entire training dataset and evaluated it on the test set, with the results shown in Table 1. It can be observed that FSF-Net generates significantly better future 3D occupancy results than Copy\&Paste, demonstrating the model's ability to learn the scene change trends. Unlike previous works that use deep learning networks alone for future scene prediction, FSF-Net employs a coarse-to-fine network structure. This approach combines the precise prediction capabilities of physical models for known voxel movement time information with the strengths of deep learning models for predicting unknown voxel semantic information, allowing for better prediction of future scene information and improving mIoU and IoU. As shown in Table \ref{tab:exp_1}, the multi-frame prediction results of our model are significantly better than other methods. For the first frame prediction, our model outperforms the previous best method \cite{occ-world} by 9.56\% in mIoU and 10.87\% in IoU.Simultaneously, we evaluated our model in the BEV space, with the results shown in Table \ref{tab:exp_2},. For the first frame prediction, our model surpasses the previous best method \cite{occ-world} by 12.1\% in mIoU and 4.07\% in IoU. Subsequently, we evaluated the quality of the model’s prediction for a single future frame on a test set of the same size, as shown in Table \ref{tab:exp_3}. Our model still achieves the best performance, exceeding the previous best method by 11.18\% in mIoU and 11.16\% in IoU. In the BEV space, our model outperforms the previous best method by 12.41\% in mIoU and 5.53\% in IoU.\par
Due to the significantly larger size of the training set compared to the validation and test sets in the Occ3D dataset, we opted to use 20\% of the training data to train the model and evaluate its generalization ability on 100\% of the evaluation set. The results are shown in Tables \ref{tab:exp_4} and \ref{tab:exp_5}. Our model’s predicted future scenes are of significantly higher quality than those of other methods. Specifically, in the evaluation of the predicted first future frame, our model outperforms the previous best method by 8.28\% in mIoU and 5.67\% in IoU. In the BEV space, compared to the previous best method, our model achieves an improvement of 7.21\% in mIoU and 2.64\% in IoU.\par

\begin{table*}[t]
\centering
\caption{End-to-end Planning Performance on nuScenes.Aux. Sup. represents auxiliary supervision apart from the ego trajectory. We use bold to denote the best results.}
\resizebox{1.0\linewidth}{!}{
\begin{tabular}{c| c c | c c c| c c c}
    \hline
    \textbf{Method}
    &  Input&Aux.Sup.
    & \multicolumn{3}{c|}{L2 (m) $\downarrow$} 
    & \multicolumn{3}{c}{Collision Rate (\%) $\downarrow$} \\
    & & & 1s & 2s & \multicolumn{1}{c|}{Avg.} 
    & 1s & 2s & Avg. \\ 
    \hline
IL & LiDAR  & None &  0.44 &1.15 & 0.80 & 0.08 & 0.27 & 0.18 \\
NMP  & LiDAR &  Box\&Motion &  0.53 & 1.25 & 0.89 & 0.04 & 0.12 & 0.08 \\
FF  &LiDAR &Freespace & 0.55 & 1.20 & 0.88 & 0.06 & 0.17 & 0.12 \\
EO &LiDAR &Freespace &  0.67 & 1.36 & 1.02 & 0.04 & 0.09 & 0.07 \\
    \hline
ST-P3 & Camera  & Map\&Box\&Depth & 1.33 & 2.11 & 1.72 & 0.23 &  0.62 & 0.43\\
UniAD  &Camera & Map\&Box\&Motion\&Tracklets\&Occ &0.48 &\textcolor{red}{\textbf{0.96}} &\textcolor{red}{\textbf{0.72}} & 0.05 & 0.17 & 0.11 \\
VAD-Tiny  & Camera  & Map\&Box\&Depth & 0.60 & 1.23 & 0.82 & 0.31 &  0.53 & 0.42 \\
VAD-Base & Camera  & Map\&Box\&Depth & 0.54 & 1.15 & 0.85 & 0.04 & 0.39 & 0.23 \\
OccNet & Camera  &3D-Occ\&Map\&Box &1.29 &2.13 &1.71 & 0.21 & 0.59 & 0.40 \\
    \hline
OccNet & 3D-Occ  &Map\&Box & 1.29 & 2.31 & 1.75 & 0.20 & 0.56 & 0.38 \\
OccWorld & 3D-Occ &  None &\textcolor{red}{\textbf{0.43}} & 1.08 & 0.76 &  0.07 & 0.38 & 0.23 \\
FSF-Net & 3D-Occ &  None & 0.54 & 1.09 & 0.82 &\textcolor{red}{\textbf{0.012}} & \textcolor{red}{\textbf{0.013}} & \textcolor{red}{\textbf{0.013}} \\
    \hline
\end{tabular}}
\label{tab:exp_6}
\end{table*}

\begin{table*}[t]
\centering
\caption{Ablation studies on Occ3d datasets. Boldfaced numbers highlight the best.}
\resizebox{1.0\linewidth}{!}{
\begin{tabular}{c| c c c c | c c | c c }
    \hline
    \textbf{Method}
    & \multicolumn{2}{c}{Miou $\uparrow$}  
    & \multicolumn{2}{c|}{Iou $\uparrow$}
    & \multicolumn{2}{c|}{L2 (m) $\downarrow$} 
    & \multicolumn{2}{c}{Collision Rate (\%) $\downarrow$} \\
    &0.5s &1s &0.5s &1s &1s & \multicolumn{1}{c|}{2s} 
    & 1s & 2s \\ 
    \hline
FSF-Net & \textcolor{red}{\textbf{42.38}} & \textcolor{red}{\textbf{28.60}} & \textcolor{red}{\textbf{48.15}} &\textcolor{red}{\textbf{36.01}} & 0.54 & 1.09 & \textcolor{red}{\textbf{0.012}} & \textcolor{red}{\textbf{0.013}} \\
w/o Quality-Fusion & 30.82 & 14.41 & 37.42 & 23.42 & 0.63 & 1.28 & 0.013 & 0.014 \\
w/o BEV Flow & 36.21 & 24.82 & 41.65 & 32.78 & 0.44 & 0.92 & 0.014 & 0.014 \\
w/o VQ-Mamba & 32.98 & 16.53 & 41.51 & 26.09 & - & - & - & - \\
\hline
VQ-Mamba  & 36.21 & 24.82 & 41.65 & 32.78 & 0.44 & 0.92 & 0.014 & 0.014 \\
w/o temporal Mamba & 31.51 & 20.05 & 38.59 & 29.73 & \textcolor{red}{\textbf{0.33}} & \textcolor{red}{\textbf{0.64}} & 0.015 & 0.014 \\
     \hline
\end{tabular}}
\label{tab:exp_7}
\end{table*}

\textbf{Visualizations.} We visualize the output results of the proposed FSF-Net in  Fig. \ref{fig:4_frame}. As shown, our method can successfully predict future scenes.Moreover, compared to the Base-Net model, our method better captures temporal information and generates more refined prediction results. Additionally, Fig. \ref{fig:2_frame}, we compare our prediction results with those of OccWorld. It can be observed that our predictions are more accurate and closer to the ground truth, especially for dynamic objects.
\subsection{Evaluation of Motion planning}
Maintaining the same model parameters configured for the 4D occupancy task, we compared our model with existing end-to-end methods in terms of planning performance, including L2 error and collision rate, using scene sequences with shorter input times. As shown in Table \ref{tab:exp_6}, FSF-Net outperforms other methods in collision rate without using maps and bounding boxes as supervision.Compared to the OccWorld method, FSF-Net reduces the collision rate by 0.058\% at the 1st second and 0.367\% at the 2nd second, due to its more refined grid occupancy maps generated from temporal information.\par
Although our model achieved the best performance in terms of collision rate, its performance in L2 error was slightly lacking. On one hand, this is because the input scene sequences of our model are shorter compared to other methods, making the extraction of temporal information from vehicle trajectories relatively coarse. On the other hand, we did not use maps and bounding boxes as supervision, relying solely on 3D occupancy and ego-vehicle trajectories as supervision. These factors make accurate predictions of future vehicle trajectories more challenging. Nevertheless, our model ranked among the top four in L2 error from smallest to largest, indicating that our approach can fairly accurately predict future ego-vehicle trajectories.
\subsection{Ablation studies}
To investigate the effectiveness of each module, we conducted more in-depth studies and designed ablation experiments to validate their contributions.\par
In this experiment, we investigated the impact of the coarse-to-fine network framework on 4D occupancy prediction and motion planning tasks. As shown in Table \ref{tab:exp_7}, where w/o Quality-Fusion refers to replacing our designed Quality-Fusion framework with a simple multi-head attention mechanism and substituting the VQ-Mamba module with a standard VQVAE structure.w/o BEV Flow denotes not using the BEV Flow module and Quality-Fusion module, relying solely on VQ-Mamba to perform 4D occupancy prediction and motion planning tasks. w/o VQ-Mamba means not using the VQ-Mamba module and Quality-Fusion module, and relying only on BEV Flow for 4D occupancy prediction and motion planning tasks. w/o temporal Mamba refers to not using Mamba to extract temporal information and performing prediction and motion planning solely with a standard VQVAE network.\par
\textbf{Coarse-to-fine framework.} As described in Sec.3, the coarse-to-fine network framework helps the network integrate the advantages of both the BEV Flow prediction module and the VQ-Mamba prediction module. As shown in Table \ref{tab:exp_7}, when using BEV Flow and VQ-Mamba separately for predictions, the results show a significant decline in 4D occupancy prediction task metrics compared to the final fused outcome. However, since our model only uses a multi-head attention mechanism to assist in predicting the ego trajectory, there is a decrease in the L2 metric for motion planning tasks. Nonetheless, our model still demonstrates improved performance in collision rate.\par
\textbf{Gating and multi-scale fusion mechanism.} In Table \ref{tab:exp_7}, we also analyze the performance of the gating and multi-scale fusion mechanisms used in the Quality-Fusion module. It can be observed that replacing the gating and multi-scale fusion mechanisms with multi-head attention leads to varying degrees of decline in Miou and Iou metrics. The experimental results indicate that the use of gating and multi-scale fusion mechanisms helps the model better integrate prediction results.\par
\textbf{The extraction of temporal information.} We evaluated the contribution of temporal information to the model's performance improvement. This study included comparing the performance when the prediction module and temporal Mamba module were removed from the VQ-Mamba network. The experimental results are shown in Table \ref{tab:exp_7}. The findings of this study reveal the contribution of temporal information to the overall network performance, particularly in terms of how it affects the accuracy of predicting future occupancy scenes.
\section{Limitations And Future Works}
\textbf{Ego trajectory prediction.} As shown in Table \ref{tab:exp_7}, our model performs poorly in predicting the ego trajectory. Although FSF-Net ultimately predicts future 3D occupancy scenes with higher quality than the coarser predictions, the ego trajectory prediction is only influenced by the generated predicted scenes through a cross-attention mechanism during the final fusion process. Since ego trajectory prediction is closely related to temporal information, we believe this limitation can be addressed by incorporating additional temporal components into the FSF-Net for ego trajectory prediction.\par
\textbf{Accurate BEV Flow.} As described in Section 3.3, due to the limitations of the input for the constrained BEV Flow module, we are unable to use the BEV Flow between the current and predicted frames to generate prediction results, which affects the accuracy of BEV Flow module predictions. Some existing research focuses on predicting optical flow between the current and future images using deep learning models. In future research, if BEV Flow for the current frame can be obtained, the accuracy issue of BEV Flow module predictions could be addressed.

\section{Conclusion}
In this work, we propose a coarse BEV scene flow 4D occupancy forecasting network FSF-Net. First, we  develop a general occupancy forecasting framework based
on coarse BEV scene flow. Second, we propose a network VQ-Mamba to enhance 4D occupancy feature representation ability. Third, we design a U-Net based quality fusion network UQF to efficiently fuse coarse occupancy maps forecasted from BEV scene flow and latent features.  Extensive experiments on the Occ3D dataset demonstrate that our method shows significant performance in forecasting the occupancy maps with semantic labels.

\section*{Acknowledgement}
This work was supported by National Key R\&D Program of China (Grant ID: 2020YFB2103501), National Natural Science Foundation of China (Grant ID: 61971203), and Major Project of Fundamental Research on Frontier Leading Technology of Jiangsu Province (Grant ID: BK20222006).

\section*{Disclosure}

The authors declare no conflicts of interest.

\appendix
\section{BEV Flow}

Inspired by the remarkable performance of optical flow in estimating pixel displacement tasks, we adopted a similar representation to depict the movement of voxels throughout the scene over time. However, calculating the movement of every voxel in the scene is inefficient and impractical due to the large number of voxels devoid of semantic information. Considering that in outdoor scenes, there are almost no voxels moving along the Z-axis and that the voxels at the same (x, y) point are often part of the same object, we made the assumption that the movement of any voxel at the same (x, y) point is identical.Therefore, by extracting the displacement of the highest voxel at each (x, y) point, we can obtain the displacement of all voxels at this point.Meanwhile, since the time interval between consecutive frames is short, we can roughly assume that for a sequence of three consecutive frames, the voxel movement between the first two frames is the same as the movement between the last two frames. Therefore, the voxel movement between the first two frames can be used to approximate the movement between the last two frames.Based on these two premises,we proposed BEV Flow module to more efficiently reflect the movement of voxels in the scene.The detailed structure of the BEV Flow module is shown in Fig. \ref{fig:module}(a).\par
First, we convert the scene S into a 2D grayscale image G of size (H, W) by finding the highest voxel at each (x, y) point. The value of each pixel in $\mathcal{B}$ corresponds to $\mathcal{S}$ The height of the voxel at the corresponding position.
First, by finding the voxel at the highest position at each (x, y) point, we convert the input scenes $\mathcal{S}^{0},\mathcal{S}^{1}$ from two consecutive frames into 2D grayscale images $\mathcal{B}_{0},\mathcal{B}_{1}$ with dimensions (H, W). The value of each pixel in $\mathcal{B}$ corresponds to the height of the voxel at the corresponding position in $\mathcal{S}$.The process of converting a voxel scene into a BEV map can be expressed as follows:
\begin{equation}
g(x, y) = \max\{z \mid (x, y, z) \in S\}
\end{equation}
where g(x,y) represents the value at the position (x,y) in the BEV map.Then, for the obtained corresponding BEV maps 
$\mathcal{B}_{0},\mathcal{B}_{1}$from two consecutive frames, the BEV Flow between the two images is derived by calculating the perspective transformation relationship between the corresponding points of $\mathcal{B}_{0},\mathcal{B}_{1}$. In this study, the perspective transformation relationship is computed using a homography matrix, as shown in the following equation.
\begin{equation}
\mathcal{B}_{0} = M\cdot\mathcal{B}_{0}
\end{equation}
\vspace{-0.5cm}
\begin{equation}
\text{Flow} = (M - 1) \mathcal{B}_{0}
\end{equation}
where M represents the homography matrix between $\mathcal{B}_{0}$ and $\mathcal{B}_{1}$,and Flow denotes the resulting BEV Flow. Finally, the input scene $\mathbf{t}_{0}$ is interpolated and deformed according to the movement of the corresponding voxels in the BEV Flow to obtain the coarse prediction result $\widetilde{\mathcal{S}}_p$ for the Scene $\mathcal{S}_p$.

\bibliography{mybibfile}

\end{document}